\def\year{2021}\relax
\documentclass[letterpaper]{article} 
\usepackage{aaai21}  
\usepackage{times}  
\usepackage{helvet} 
\usepackage{courier}  
\usepackage[hyphens]{url}  
\usepackage{graphicx} 
\usepackage{natbib}  
\usepackage{caption} 
\usepackage{graphicx}
\usepackage{comment}
\usepackage{amsmath,amssymb} 
\usepackage{color}

\usepackage[table]{xcolor}

\usepackage[utf8]{inputenc} 
\usepackage[T1]{fontenc}    
\usepackage{hyperref}       
\usepackage{url}            
\usepackage{booktabs}       
\usepackage{amsfonts}       
\usepackage{nicefrac}       
\usepackage{microtype}      
\usepackage{adjustbox}
\usepackage{amsthm}
\usepackage{multirow}
\usepackage{amsmath}
\usepackage{subcaption}
\usepackage[ruled]{algorithm2e}

\newcommand{\mycomment}[1]{}

\usepackage{lipsum}
\usepackage{enumitem}
\usepackage{wrapfig}

\newcommand{\x}{\mathbf{x}}
\newcommand{\xnom}{\mathbf{x}_{nom}}
\newcommand{\z}{\mathbf{z}}
\newcommand{\znom}{\mathbf{z}_{nom}}
\newcommand{\W}{\mathbf{W}}
\newcommand{\bias}{\mathbf{b}}
\newcommand{\real}{\mathbb{R}}
\newcommand{\eps}{\epsilon}
\newcommand{\bu}{\mathbf{u}}
\newcommand{\bl}{\mathbf{l}}
\newcommand{\A}{\mathbf{A}}
\newcommand{\reg}{\mathcal{R}}
\newcommand{\loss}{\mathcal{L}}
\newcommand{\lam}{\lambda}
\newcommand{\bv}{\mathbf{v}}

\newcommand{\E}{\mathbb{E}}

\newcommand{\singleprop}{{\textsf{\small SingleProp}}\xspace}
\newcommand{\singlepropfastlin}{{\textsf{\small SingleProp-FastLin}}\xspace}
\newcommand{\singlepropzero}{{\textsf{\small SingleProp-Zero}}\xspace}

\newcommand{\Lily}[1]{\textcolor{orange}{[Lily: #1]}}
\newcommand{\SL}[1]{\textcolor{blue}{SL:#1}}

\title{Fast Training of Provably Robust Neural Networks by \textit{SingleProp}
}
\author {
    Akhilan Boopathy\textsuperscript{\rm 1},
    Tsui-Wei Weng \textsuperscript{\rm 1},
    Sijia Liu\textsuperscript{\rm 2},
    Pin-Yu Chen\textsuperscript{\rm 2},
    Gaoyuan Zhang\textsuperscript{\rm 2},
    Luca Daniel\textsuperscript{\rm 1}
    
}

\affiliations {
    \textsuperscript{\rm 1} Massachusetts Institute of Technology \\
    \textsuperscript{\rm 2} MIT-IBM Watson AI Lab, IBM Research
}

\begin{document}

\maketitle

\begin{abstract}

Recent works have developed several methods of defending neural networks against adversarial attacks with certified guarantees. However, these techniques can be computationally costly due to the use of certification during training. We develop a new regularizer that is both more efficient than existing certified defenses, requiring only one additional forward propagation through a network, and can be used to train networks with similar certified accuracy. Through experiments on MNIST and CIFAR-10 we demonstrate improvements in training speed and comparable certified accuracy compared to state-of-the-art certified defenses.
\end{abstract}

\mycomment{
\Lily{If any of the following works, will need to add into the current version
\begin{itemize}
    \item save the best model: any improvement
    \item trades loss, schedule: any improvement
    \item singleprop verifier: does the new certified accuracy improve?
\end{itemize}
}}

\section{Introduction}


Although deep neural networks (DNNs) have achieved tremendous success in various applications, it has become widely-known that they are vulnerable to adversarial examples (also known as adversarial attacks), namely, crafted  examples with human-imperceptible perturbations to cause  misclassification \citep{goodfellow2014explaining,szegedy2013intriguing}. 
Many  attack generation methods have been proposed in order to find the possible minimum adversarial perturbation, commonly evaluated by its $\ell_p$ norm for $p \in \{ 0, 1,2,\infty  \}$ \citep{papernot2016limitations,carlini2017towards,athalye2017synthesizing,su2019one,xu2018structured,chen2017ead}.  Meanwhile, various defense methods  were proposed to enhance the robustness of DNNs against adversarial attacks. However, many of them are built on heuristic strategies, which are thus easily bypassed by  stronger adversaries \citep{athalye2018obfuscated}.
The work \cite{madry2017towards} proposed a stronger defense method, adversarial training, which  minimizes the \textit{worst-case} training loss under adversarial perturbations.

Motivated by the limitation of heuristic defense, another line of research (known as verified/certified robustness) aims to
provide \textit{provable} robustness guarantees of DNNs against an input with \textit{arbitrary perturbation} within a certain $\ell_p$ ball region \citep{katz2017reluplex,cheng2017maximum,carlini2017ground,kolter2017provable,raghunathan2018certified,weng2018towards,zhang2018crown,Boopathy2019cnncert,dvijotham2018dual,wong2018scaling,xiao2019training,gowal2019effectiveness,mirman2018differentiable,dvijotham2018training}.  
The recent progress on verification   spans from the exact verification method \citep{katz2017reluplex,cheng2017maximum,carlini2017ground} to the relaxed verification method \citep{kolter2017provable,raghunathan2018certified,weng2018towards,singh2019abstract,zhang2018crown,Boopathy2019cnncert,dvijotham2018dual}. Here the former uses expensive computation methods, e.g., mixed-integer programming (MIP), to find the exact minimum adversarial perturbation, and the latter considers a relaxed verification problem by convexifying the adversarial polytope which significantly improves the computation efficiency compared to the exact method at the cost of tightness of robustness certificate.

Another research direction that has attracted a lot of interest in this field is robust training, i.e. to make a neural network classifier more robust to adversarial attacks. Recent work \cite{xiao2019training} proposed the principle of co-design between training and verification, and showed that the exact verification method~\cite{tjeng2019evaluating} can be accelerated by imposing weight sparsity and activation stability (so-called ReLU stability) on trainable neural network models. On the other hand, there are several works~\cite{kolter2017provable,raghunathan2018certified,gowal2019effectiveness,mirman2018differentiable} aiming to train a more \textit{verifiable} model targeted on different verifiers mentioned earlier~\cite{kolter2017provable,raghunathan2018certified,weng2018towards,singh2019abstract,zhang2018crown,Boopathy2019cnncert,wong2018scaling} and this line of research is known as \textit{certified robust training}. The idea is to incorporate the robustness verification bounds into the training process and thus the learnt model yields strengthened robustness with certificate. Nevertheless, current verification-based training methods are multiple times slower than standard (non-robust) training per training step. In particular, using convex outer bounds-based methods~\cite{kolter2017provable,weng2018towards,zhang2018crown} empirically require more than $100\times$ standard training cost~\cite{kolter2017provable}. The fastest method to date, interval bound propagation (IBP)~\cite{gowal2019effectiveness}, requires 2 additional forward propagations compared to standard training, for a total training time of $3 \times$ standard training. We highlight that IBP is significantly faster than adversarial training, which typically requires much greater than 2 adversarial steps to achieve high robustness~\cite{madry2017towards}. As such, IBP is currently the fastest effective certified robust training method.

While IBP empirically achieves high certified accuracies, i.e. higher percentage of the test images that are guaranteed to be classified correctly under any possible $\ell_p$ perturbations with magnitude $\epsilon$, it still requires $2 \times$ additional computation overhead of standard training. This raises the question of what minimum computational overhead is required to achieve certified robustness for a neural network model, which is the main motivation of this work. Our goal is to develop a robust training method that achieves high certified accuracy while achieving the minimum overhead of~\textit{only one additional} forward pass than standard training. We summarize our contributions as follows: 
\begin{itemize}
    \item We propose an efficient robust training algorithm, named as \singleprop, based on a novel regularizer which can be derived as an approximation of linear bounding verifiers~\cite{kolter2017provable,weng2018towards,zhang2018crown,Boopathy2019cnncert}. Our proposed regularizer requires only 1 additional forward pass per training step (hence the name \singleprop) relative to standard training, resulting in only $1 \times$ additional training time and memory overhead relative to standard training. To our best knowledge, \textbf{\singleprop} is the fastest and most efficient among SOTA certified training algorithms.  
    
    
    \item Extensive experiments demonstrate \singleprop achieves superior computational efficiency and comparable certified accuracies compared to the current fastest certified robust training method IBP~\cite{gowal2019effectiveness}. In particular, we show on both MNIST and CIFAR datasets that the certified accuracies only decrease slightly while enjoying $1.5\times$-$2\times$ faster to train as well as $1.5 \times$ reduction in memory usage. While this drop in accuracy is expected due to the approximation used in our method, we observe that \singleprop outperforms IBP training on the Fast-Lin verifier~\cite{weng2018towards}.\mycomment{in certain cases our methods outperform IBP. \Lily{what are the cases outperform?}}

\mycomment{
    \item \Lily{If the singleprop verifier or the new loss works, then we have another contribution here.}}
    
    
    
    \mycomment{
    \item We extend our regularizer to an $\ell_0$ threat model and model ensembles. We demonstrate empirically that our regularizer achieves high certified accuracy on this threat model. We also empirically show that model ensembling of models trained with our regularizer achieve higher certified accuracies.}  
\end{itemize}

\section{Background and Related Work}
\label{sec:background}

\subsection{Verifications}

Assuming a norm-bounded threat model, finding the minimum adversarial distortion exactly is an NP-complete problem, making it computationally infeasible~\citep{katz2017reluplex}. Fortunately, finding lower bounds on the minimum adversarial distortion is computationally tractable. Several techniques find these lower bounds only as a function of model weights ~\citep{szegedy2013intriguing,peck2017lower,hein2017formal,raghunathan2018certified}, but these methods typically provide very loose bounds for neural networks with more than 2 layers. Using an input-specific certification method, it is possible to find non-trivial bounds for fully connected ReLU networks~\citep{kolter2017provable,weng2018towards,wang2018efficient}, as well as networks with general activation functions~\citep{zhang2018crown} and general CNN and RNN architectures~\citep{Boopathy2019cnncert, ko2019popqorn}.

\subsection{Certified Defenses}
Recent works have also developed methods of defending against adversarial attacks. One line of work uses adversarial training with adversarial attacks and empirically demonstrates high resistance to attacks~\citep{madry2017towards,sinha2017certifiable}. However, adversarial training is not targeted towards verification or certification methods and is therefore not considered as a certified defense. One step towards certified defenses is natural regularizations on the model parameters, such as sparsity-inducing weight magnitude penalization. This method combined with adversarial training yields highly verifiable models~\citep{xiao2019training}. Using an additional ReLU stability regularizer to enhance ease-of-verification allows for even more verifiable models, but at the cost of at least $3\times$ of standard training even without adversarial training. Since layer-specific regularizations alone empirically do not help certifiable robustness, in this paper we focus on the minimum computational overhead achievable using additional forward passes. 

Other defenses specifically target certifiers or use certification methods as part of the training procedure, and this type of defenses is known as \textit{certified defense}. We note that these ``certified'' defenses are not truly certified since they cannot ensure robustness to unseen points without using a certifier on these points. These defenses instead produce models that are empirically more certifiable on unseen test points. Using convex outer bounds to bound the adversarial loss function has been shown to be effective at producing more certifiable models on the targeted verifiers~\cite{kolter2017provable,weng2018towards}, although training is relatively slow~\citep{kolter2017provable,wong2018scaling}. Using interval bounds propagation (IBP) to bound the adversarial loss is much cheaper to train~\citep{gowal2019effectiveness} and has surprisingly become the state-of-the-art certifiably robust training method~\citep{gowal2019effectiveness,salman2019convex} despite IBP generally performing much worse than the convex outer bounds~\citep{kolter2017provable,wong2018scaling} in certifications of standard networks. Recently, another verifier was developed by combining IBP with the CROWN certifier~\cite{zhang2018crown} to produce CROWN-IBP training. The authors show that the trained networks could outperform IBP in certified accuracy by up to $2.74 \%$ on MNIST and up to $9.16 \%$ on CIFAR \mycomment{\Lily{Akhilan summarize the improved acc of crown-ibp on mnist and cifar}}. However, CROWN-IBP is much more computationally expensive during training, specifically $9 \times$ slower than IBP~\cite{zhang2020crownibp}, which is equivalent to imposing total additional $26 \times$ cost of standard training. As our emphasis is on the \textit{efficiency} of certifiable robust training, we focus on comparing with IBP rather than CROWN-IBP. In summary, we develop an efficient certified defense that has even lower computation overhead (only $1 \times$ additional to standard training cost) than IBP-based defenses~\cite{gowal2019effectiveness, zhang2020crownibp} while yielding comparable performance with existing methods.

\paragraph{Threat model.}
In this paper, we will use the notation of fully-connected neural networks for exposition, but our method works for general convolutional neural networks including residual networks. Appendix~\ref{app:notation} Table~\ref{tab:Notation_updated} includes
\mycomment{\SL{summarize the main notations...?} \Lily{Akhilan moved notation table here with updated table.}} descriptions of the main notations used in this paper. Consider an $n$ layer neural network $f(\x)$ with input $\x$ where the first layer of the network $\z^0$ is set to $\x$. Given weights $\W^i$, biases $\bias^i$ and an activation function $\sigma$,  for $i = 0,\ldots, n-1$, subsequent layers are defined as:
\begin{equation}\label{eqn:z_def}
    \z^{i+1} = \W^{i+1} \sigma(\z^i)+\bias^{i+1},
\end{equation}
with $f(\x)=\z^n$. With this expression, the first layer of the network is defined by using the identity activation at the first layer. We assume the following threat model: a nominal input $\xnom$ is perturbed by perturbation $\delta$ to produce a perturbed input $\x = \xnom + \delta$, where $||\delta||_p \leq \eps$ and  $||\cdot||_p$ represents an $\ell_p$ norm. Suppose the correct classification is given by $c$. Then the minimum distortion $\eps^*$ for misclassification is the minimal $\eps \in \real^+$ satisfying: $\max_{j \neq c} \mathbf{z_j^n} - \mathbf{z_c^n} > 0$.

\paragraph{Interval Bounds Propagation.} There exist several methods to efficiently find certified lower bounds on the minimum distortion necessary for misclassification. One such method is interval bounds propagation (IBP) ~\citep{gowal2019effectiveness,Gehr2018AI2} which bounds each layer in a network with a fixed upper and lower bound. These bounds are then propagated at each layer of the network using the previous layer's bounds. Specifically, given layer-wise bounds where $\bl^i \leq \mathbf{z^i} \leq \bu^i$, the next layer's bounds are found as:
\begin{equation}\label{eq: ibp_u}
    \bu^{i+1} = \W^{i+1}_+ \sigma(\bu^i) + \W^{i+1}_- \sigma(\bl^i) + \bias^{i+1},
\end{equation}
where $\W_+$ and $\W_-$ denote the positive and negative components of $\W$ respectively with other entries being zeros otherwise. Lower bounds are found similarly. Intuitively, IBP finds a box bounding each layer, which can result in very loose bounds for general network as demonstrated in ~\citep{kolter2017provable,Gehr2018AI2}.

\paragraph{Linear Bounding Framework.} Certified bounds can also be found using a linear bounding framework as first proposed in Fast-Lin ~\citep{weng2018towards} and later in the Neurify ~\citep{wang2018efficient} and DeepZ ~\citep{singh2018fast} frameworks. This approach typically finds tighter bounds on minimum distortion than IBP. This framework bounds each activation layer $\sigma(\z^i)$ as follows: $\alpha_L^i \odot \z^i + \beta_L^i \leq \sigma(\z^i) \leq \alpha_U^i \odot \z^i+ \beta_U^i$,
where $\alpha_L^i, \alpha_U^i$ represents the slopes of linear bounds on the activation and $\beta_L^i$,$\beta_U^i$ represents intercepts of linear bounds on the activation. When $\sigma$ is the ReLU activation, provided bounds $\z_L^i, \z_U^i$ on $\z^i$ satisfying $\z_{L}^i \leq \z^i \leq \z_{U}^i$, Fast-Lin sets the coefficients to be: \mycomment{\Lily{akhilan use general $\alpha^i_{U,j}, \alpha^i_{L,j}$}}
\begin{equation*}
    \alpha^i_{L,j} = \alpha^i_{U,j} = \frac{\z_{U,j}^i}{\z_{U,j}^i-\z_{L,j}^i}, \; \beta_{L,j}^i = 0, \; \beta_{U,j}^i = -\frac{\z_{U,j}^i \z_{L,j}^i}{\z_{U,j}^i-\z_{L,j}^i},
\end{equation*}
if the neuron $j$ is uncertain, meaning $\z_{L,j}^i < 0$, $\z_{U,j}^i > 0$. When both bounds are positive or negative, the bound on the activation is exactly the linear component on the corresponding side (i.e. when $\z_{L,j}^i > 0$ for example, $\alpha^i_{L,j} = \alpha^i_{U,j} =1$, $\beta_{L,j}^i=\beta_{U,j}^i=0$). Using these layer-wise bounds, Fast-Lin finds a pair of linear bounds on the network: $\A_L \x + \bias_L \leq f(\x) \leq \A_U \x + \bias_U$. Then Fast-Lin bounds the network output over all possible adversarial distortions measured by $\epsilon$-$\ell_p$ ball by:
\begin{equation}
    \A_L \xnom + \bias_L - \eps ||\A_L||_{:,q} \leq f(\x) \leq \A_U \xnom + \bias_U + \epsilon ||\A_U||_{:,q},
\end{equation}
\mycomment{\Lily{Didn't define $\z_L,\z_U$ here}}
where $||\cdot||_{:,q}$ denotes a row-wise $q$ norm, dual to the norm $p$ of the assumed attack threat model. $\epsilon$ is the assumed attack norm size. Intuitively, Fast-Lin finds linear upper and lower bounds on the entire network to analyze the output layer. Because Fast-Lin finds linear bounds on the network as an intermediate step to finding output bounds $\z_L,\z_U$, the bounds are tighter than the corresponding IBP bounds $\bu,\bl$. Fast-Lin is equivalent to using convex outer bounds to bound the set of possible values at each layer of the network. Fast-Lin has been extended to general activation functions and asymmetric upper and lower bounds with different values of $\alpha_L^i, \alpha_U^i$ in CROWN~\citep{zhang2018crown}, and has been extended to general network architectures in CNN-Cert~\citep{Boopathy2019cnncert}.


\section{\textsf{SingleProp}: An Efficient Robust Training Framework}
\label{sec:method}
In this section, we propose a new robust training method $\singleprop$ which is $1.5 \times$ more computationally efficient than the most efficient SOTA certified training algorithm. We start by first deriving $\singleprop$ regularizers as approximations of linear bounding certifiers in Sec~\ref{sec:regularizer}. We then analyse the run time of our method in Sec~\ref{sec:runtime} and detail the training procedure in Sec~\ref{sec:training_detail}. \mycomment{\Lily{Need to enable the section numbers. Here the sec number of training procedure disappeared.}}
\mycomment{\Lily{In addition to the computation time, we can also add how much less memory required for computation, and add this number into contribution in the Introduction number.}}

\subsection{Robust loss function with \singleprop regularizers}
\label{sec:regularizer}
\mycomment{\Lily{The $L$ shoud be $\mathcal{L}$ in the equations?} \textcolor{blue}{Akhilan: $\mathcal{L}$ is a function of $\theta$, while $L$ is a function of the network output}}
Let $\theta$ denotes the parameters of neural network $f_\theta$, and let $\loss_\theta(\z^n, \mathbf{y})$ be a standard loss function as a function of network output $\z^n$ and one-hot-encoded label $\mathbf{y}$. Many robust training methods including IBP can all be interpreted as adding a regularizer $\reg_\theta(\bl^n, \bu^n, \mathbf{y})$ to the standard loss function and thus forming a \textit{robust} loss function: $\loss_\theta(\z^n, \mathbf{y}) + \lam \reg_\theta(\bl^n, \bu^n, \mathbf{y})$, where $\bu^n$ and $\bl^n$ are layer-wise output bounds of the neural network $f_\theta(\x)$ when the input $\x$ is perturbed and $\lam$ is a regularization parameter. The IBP regularizer can be written in the following form~\cite{gowal2019effectiveness}: 
\begin{equation}\label{eq:ibp_regularizer}
    \reg_\theta(\bl^n, \bu^n, \mathbf{y}) = \E[\mathcal{L}(\bl^n\circ \mathbf{y} + \bu^n\circ (1-\mathbf{y}),\mathbf{y}) - \mathcal{L}(\znom^n,\mathbf{y})],
\end{equation}
where  $\znom^n=f_\theta(\xnom)$ is the output at unperturbed input $\xnom$ \mycomment{\Lily{akhilan finishes... when input is perturbed ..., and $y$ is not defined..., $\znom^n$ is not defined too. Also, is $L$ to be in caligraphic form?}} and $\circ$ denotes element-wise multiplication. However, IBP requires two additional propagations through the network during training relative to standard training due to computing $\bu^n$ and $\bl^n$. Instead of using two quantities $\bu^n$ and $\bl^n$ to propagate uncertainty, we design a new regularizer:
\begin{align}
    & \reg_\theta(\bv^n,\znom^n,\mathbf{y}) = \E[\mathcal{L}((\znom^n-\bv^n)\circ \mathbf{y} \label{eq:single_margin_regularizer_v2} \\ \nonumber
    & \quad \quad \quad + (\znom^n+\bv^n)\circ (1-\mathbf{y}),\mathbf{y}) - \mathcal{L}(\znom^n,\mathbf{y})]
\end{align}
where $\bv^n$ is a trainable quantity that can be computed with only a \textit{single propagation} and is thus more efficient than the IBP regularizer. In fact, the quantity $\bv^n$ is motivated by approximating the bounds in linear bounding certifiers such as Fast-Lin~\cite{weng2018towards}. In the following, we show that it is possible to compute $\bv^n$ with only a \textit{single propagation} in each training step with the derived recursive relation Eq.~\eqref{eqn:v_def}. 
 
\paragraph{Derivation of \singleprop regularizer.} We use fully-connected network for easier exposition, but our method can be easily extended to CNNs or residual networks. As described in Sec~\ref{sec:background}, the output of NN can be bounded by a lower bound $\z_L = \A_L \xnom + \bias_L - \eps ||\A_L||_{:,1}$ and an upper bound $\z_U = \A_U \xnom + \bias_U + \eps ||\A_U||_{:,1}$ in the linear bounding method assuming an $\ell_\infty$ perturbation norm\footnote{other $p$ can be derived similarly.}. At each layer $i$, we wish to approximate the bounds $\z^i_L$ and $\z^i_U$ using only a single forward propagation. To do so, we define the half bound gap at layer $i$ as $\bv^i = \frac{1}{2} (\z^i_U - \z^i_L)$. At all layers $i$, the dimension of each $\bv^i$ is the same as the corresponding layer $\z^i$. We will show that for $i>0$, $\bv^i$ can be approximated recursively using only the previous value $\bv^{i-1}$. To avoid the need for additional forward propagations, we will make an additional assumption on the \textit{average} of the bounds $\frac{1}{2}(\z_U + \z_L)$. This will allow us to approximate the range of values each layer $\z$ can take as $[\znom-\bv,\znom+\bv]$, where $\znom$ represents the value of the layer for unperturbed input $\xnom$. To derive the recursive approximation of $\bv$, we first start with the definition of $\bv^i$ in terms of $\z^i_L$ and $\z^i_U$ in (i) below. Expanding $\z^i_L$ and $\z^i_U$ by their definitions implies (ii). The terms involving $||\cdot||_{:,1}$ can be upper bounded to yield (iii) using the fact that $||AB||_{:,1} \leq |A| ||B||_{:,1}$ elementwise. This can be expressed recursively in terms of $\bv^{i-1}$ in (iv). Finally, using the assumption that the average of the bounds is approximately the value of the network at $\xnom$ (i.e. $\znom^{i-1} \approx \frac{1}{2}[\z^{i-1}_U + \z^{i-1}_L]$), this reduces to (*):
\begin{flalign}
    \bv^i &\underset{(\text{i})}=  \frac{1}{2} (\z^i_U - \z^i_L)  \nonumber \\
    &\underset{(\text{ii})}=
    \frac{1}{2} [|\W^i| (\alpha_U^{i-1}(\z^{i-1}_U - \eps ||\A^{i-1}_U||_{:,1})+\beta_U^{i-1} ) \nonumber  \\ &~~~~ -|\W^i| (\alpha_L^{i-1}(\z^{i-1}_L + \eps ||\A^{i-1}_L||_{:,1})+\beta_L^{i-1})] \nonumber \\
    &~~~~ + \frac{1}{2}[\eps ||\A^i_U||_{:,1} +  \eps ||\A^i_L||_{:,1}] \nonumber  \\
    &\underset{(\text{iii})}\leq \frac{1}{2} [|\W^i| (\alpha_U^{i-1} \z^{i-1}_U+\beta_U^{i-1} ) - |\W^i| (\alpha_L^{i-1}\z^{i-1}_L+\beta_L^{i-1})] \nonumber \\
    &\underset{(\text{iv})}= |\W^i| \frac{\alpha_U^{i-1}+\alpha_L^{i-1}}{2} \bv^{i-1} + \frac{1}{2} |\W^i| (\beta_{U}^{i-1}-\beta_{L}^{i-1}) \nonumber \\ 
    &~~~~ +|\W^i| \frac{\alpha_U^{i-1}-\alpha_L^{i-1}}{4} (\z^{i-1}_U + \z^{i-1}_L) \nonumber  \\
    &\underset{(*)}\approx |\W^i| [\frac{\alpha_U^{i-1}+\alpha_L^{i-1}}{2} \bv^{i-1} + \frac{\beta_{U}^{i-1}-\beta_{L}^{i-1}}{2} \nonumber\\ &~~~~  +  \frac{\alpha_U^{i-1}-\alpha_L^{i-1}}{2} \znom^{i-1}] \label{eqn:v_def}
\end{flalign}
Eq.~\eqref{eqn:v_def} with equality defines the propagation of quantity $\bv$ through the network. Given $\bv^0$, this provides an update equation to find $\bv$ for subsequent layers, which can be used to approximate bound margins at all layers. Since $\bv^0$ corresponds to bounds at the first layer, $\bv^0$ is initialized to $\eps \mathbf{1}$. We note that Eq.~\eqref{eqn:v_def} actually holds as an approximate inequality ($\lesssim$), and by treating it as an equality ($=$) in our computation of $\bv$, we approximately overestimate the true bounds of the linear bounding certifier. Despite being only an approximate overestimation of true bounds, using $\bv$ during training is justifiable because empirically the approximation is highly accurate: averaging $z_U, z_L$ is very close to $z_{nom}$ with a difference is on the order of 1e-7 (see experiments in Sec~\ref{sec: approx_error}).


\noindent Note that the specific values of $\alpha^{i-1}$ and $\beta^{i-1}$ in Eq.~\eqref{eqn:v_def} will depend on the exact bounds used for the activation function considered. Different bounds on the activation function correspond to approximating different certifiers. In the case of ReLU, the quantities in the expression above depend on exact values of $(\znom^{i-1}-\bv^{i-1})_j$ and $(\znom^{i-1}+\bv^{i-1})_j$ for each neuron $j$. In other words, the bracketed quantity in Eq.~\eqref{eqn:v_def} related to neuron $j$ can be written as:
\begin{equation*}
    \frac{\alpha_{U,j}^{i-1}+\alpha_{L,j}^{i-1}}{2} \bv^{i-1}_j + \frac{\beta_{U,j}^{i-1}-\beta_{L,j}^{i-1}}{2} + \frac{\alpha_{U,j}^{i-1}-\alpha_{L,j}^{i-1}}{2} (\znom^{i-1})_j, 
\end{equation*}
which is equivalent to the following equations depending on the neuron status:
\begin{equation}
\label{eq:singleprop_fastlin}
\begin{cases}
  \alpha_{j}^{i-1} \bv^{i-1}_j & \text{, if neuron $j$ is stable} \\
  \frac{3}{4} \bv^{i-1}_j + \frac{1}{2} (\znom^{i-1})_j - \frac{(\znom^{i-1})_j^2}{4 \bv^{i-1}_j} & \text{, if neuron $j$ is unstable}
\end{cases}
\end{equation}
We refer a neuron to be stable if it satisfies $(\znom^{i-1}+\bv^{i-1})_j (\znom^{i-1}-\bv^{i-1})_j >0$, and the resulting upper and lower bounds on ReLU are equal, where $\alpha_{j}^{i-1}$ is $1$ for ReLUs with positive inputs and $0$ for ReLUs with negative inputs. On the other hand, we refer a neuron to be unstable if $(\znom^{i-1}+\bv^{i-1})_j >0$ and $(\znom^{i-1}-\bv^{i-1})_j < 0$), and the bracket quantity in Eq.~\eqref{eqn:v_def} can be re-written as in Eq.~\eqref{eq:singleprop_fastlin}, which we call this choice of activation bounds \singlepropfastlin. For methods such as CROWN~\cite{zhang2018crown} which use adaptive selection of lower bounds on ReLU, the value of expression for unstable neurons depends on the choice of lower bound slope $\alpha_{L,j}^{i-1}$. We consider the case where unstable neurons have a lower bound of slope 0 and we call this choice of activation bounds~\singlepropzero, which yields the following equations:
\begin{equation}
\label{eq:singleprop_zero}
\begin{cases}
  \alpha_{j}^{i-1} \bv^{i-1}_j & \text{, if neuron $j$ is stable} \\
  \frac{1}{2} (\znom^{i-1} + \bv^{i-1})_j & \text{, if neuron $j$ is unstable}
\end{cases}
\end{equation}
Note that these bounds correspond to a variation of CROWN~\cite{zhang2018crown} where the lower bound is always chosen to have slope zero, which is a strictly stronger verifier than IBP since IBP can be seen as a variation of CROWN with constant upper and lower bounds. Therefore, \singlepropzero can be seen as approximating IBP bounds with a single additional propagation compared to two for IBP.


It is worth noting that deriving the true upper bound on the adversarial loss (as done by existing certified defenses~\cite{kolter2017provable,raghunathan2018certified,gowal2019effectiveness} is not necessary, since they cannot ensure robustness to unseen points without using a certifier on these points. Hence, the success of \singleprop on achieving better efficiency is due to the use of the approximated upper bound on the adversarial loss, and the effectiveness of \singleprop is demonstrated in the our experiments. We are also not aware of any exact certification-based regularizer that is competitive with our method in terms of computational efficiency: we use only $2\times$ the memory and training time of standard training (vs. at least $3-27\times$ for other methods). \mycomment{\Lily{highlight speed and memory improvement}}


\mycomment{\Lily{Can we make Algorithm on top?}}

\subsection{Run time analysis} \mycomment{\Lily{Add memory usage here, compare both IBP and Crown-ibp and highlight our improvement}}
\label{sec:runtime}
The robust loss $\mathcal{L}_\theta + \lam \reg_\theta$ requires computation of both the last layer of the unperturbed network $\znom^n$ and $\bv^n$. Note that given the unperturbed value of all intermediate layers $\znom^i$, $\bv^i$ for all layers can be computed with a single forward pass using~\eqref{eqn:v_def}. Since the unperturbed layers can be computed with a single forward pass, computing the robust loss requires two forward passes total.

During training, computing gradients of the regularized loss with respect to network parameters requires computing the gradients with respect to intermediate layers which we denote as $\nabla{ \znom^i}$ and $\nabla{ \bv^i}$ for all $i$. Note that by Equations~\eqref{eqn:z_def}~and~\eqref{eqn:v_def}, $\nabla{ \znom^i}$ can be computed using the value of the next layer's gradients $\nabla{ \znom^{i+1}}$ and $\nabla{ \bv^{i+1}}$. Similarly, $\nabla{ \bv^i}$ can be computed from $\nabla{ \bv^{i+1}}$. This implies that $\nabla{ \bv^i}$ for all layers can be computed with a single backward propagation. These gradients can be used to compute $\nabla{ \znom^{i+1}}$ for all layers with a single additional backward propagation, resulting in two backward propagations total. In summary, $\singleprop$ requires only one additional forward pass and one additional backward pass relative to standard training. Therefore, assuming that numerical operations of fixed dimensionality are performed constant time and treating as negligible the cost of layer-wise operations such as activation functions, $\singleprop$ training only requires $1$ additional forward pass and $1 \times$ of additional memory overhead compared to standard training, which is $1.5 \times$ faster than IBP~\cite{gowal2019effectiveness} (empirically $1.5$-$2\times$ faster) in speed and $1.5 \times$ reduction in the memory usage. To our best knowledge, \textbf{\singleprop} is the fastest and most efficient certified training procedures.  





\subsection{Training Procedure}
\label{sec:training_detail}

Training proceeds by using standard optimizers on the robust loss. See Appendix~\ref{app:algo} \mycomment{\Lily{number does not appear}}Algorithm~\ref{algo:singlemargin_algo} for the full procedure. Note that using a value of $\lam=0$ corresponds to standard training while using $\lam>0$ corresponds to using the regularizer \mycomment{\SL{[Is this true? not only reg. right?]}} (i.e. robust loss). In practice, robust training methods such as IBP typically increase the value of $\lam$ during training process. It is also worth mentioning that the value of $\lam$ is separate from the value of $\eps$ used during training which parameterizes the regularizer $\reg_\theta$. A value of $\eps=0$  corresponds to regular training, and higher values of $\eps$ correspond to defending against larger perturbation attacks. In addition to increasing $\lam$ during training, robust training methods also typically increase $\eps$ during training.
\paragraph{Adaptive Hyperparameter Selection} The exact schedules of $\lam$ and $\eps$ represent a large hyperparameter space and can require careful tuning for methods like IBP~\cite{gowal2019effectiveness}. To ensure consistent performance without extensive hyperparameter tuning, we propose an adaptive method of selecting the regularization hyperparameter $\lam$ using a validation set. Specifically, at each epoch, we set the value of $\lam$ as: $\lambda = \frac{\gamma \loss_{val,\theta}(\z^n, \mathbf{y})}{(1+\gamma)\loss_{val,\theta}(\z^n, \mathbf{y}) + \reg_{val,\theta}(\cdot,\cdot, \mathbf{y})}$,
where $\loss_{val,\theta}$ and $\reg_{val,\theta}$ represent quantities computed on the validation set and $\gamma>0$ is a constant hyperparameter. Intuitively, this choice of $\lam$ ensures that standard accuracy is maintained throughout the training process. If standard performance is high relative to robust performance, $\lam$ will be low and vice versa. Therefore, as robust performance increases during the course of training, $\lam$ will gradually increase. The parameter $\gamma$ controls the trade-off between robustness and standard accuracy, with smaller choices of $\gamma$ corresponding to a higher preference for standard accuracy. As a baseline, we also use a piece-wise linear schedule for $\lambda$ starting from $0$ and increasing to $0.5$, following the parameters used by IBP~\cite{gowal2019effectiveness}. We also use the piece-wise linear $\eps$ schedule suggested by IBP, where $\eps$ is set to $0$ for a warm-up period, followed by a linear increase until reaching the desired value, after which $\eps$ stays at this value. The schedule of learning rates is tuned for each method individually using a validation set.

\section{Experiments}
\mycomment{
\Lily{add the results from rebuttal and follow the reviewers suggestion:
\begin{itemize}
    \item compare with additional baselines: standard training, adv training and pure TRADES
    \item probably want to move the result of IBP+SingleProp part to another table to just show there are complemenraryness between IBP and SingeProp, and extract the relevant text in this section to a single paragraph discussion. Can put the discussion we wrote in the rebuttal about this in that paragraph too. 
    \item organize the tables, current tables are scattering around. 
\end{itemize}}}
\paragraph{Implementation, Architectures, Training Parameters.}
We directly use the code provided for IBP~\cite{gowal2019effectiveness} and use the same CNN architectures (small, medium, large and wide) on MNIST and CIFAR-10 datasets. We use adaptive hyperparameter selection for $\lambda$ or a piecewise linear schedule for $\lambda$ for all robust training methods\mycomment{\Lily{Is it true? we also have linear?}}. The MNIST networks are trained for 100 epochs each with a batch size of 100 while the CIFAR networks are trained for 350 epochs each with a batch size of 50. We use the standard values of $\eps = 0.3$ for MNIST and $\eps =  8/255$ as the training target perturbation size $\eps_{train}$. Following~\cite{gowal2019effectiveness}, the schedule of $\eps$ starts at $0$ for a warmup period ($2000$ training steps on MNIST, $5000$ training steps on CIFAR), followed by a linear increase to the desired target perturbation size ($10000$ training steps on MNIST, $50000$ training steps on CIFAR), after which $\eps$ is fixed at the target level. Additional details are reported in Appendix~\ref{app:impl}.


\paragraph{Comparative Methods \& Evaluation metric.} 
We compare the two variants of our method \singlepropfastlin and \singlepropzero with the baseline IBP~\cite{gowal2019effectiveness} on the same metric, i.e. certified accuracy (acc.). We also report the total combined fraction of points certifiable by \textit{either} networks trained with IBP or \singleprop which we call IBP+\singleprop, which will always be greater or equal than the certified acc. by individual methods since it finds the union of points certifiable under individual IBP or SingleProp models. This \textbf{\textit{does not}} correspond to certified accuracy on a robustly trained model or model ensemble, but is included to show the complementarity between the certifications of IBP and SingleProp models.
The main verifier used to compute certified accs. is IBP, which will favor IBP-trained models. However, we show that \singleprop not only has competitive certified accs. on the IBP verifier, it actually outperforms IBP-trained models on the Fast-Lin verifier~\cite{weng2018towards}. On the IBP verifier, we report the full test set results (10000 points) over a range of perturbation sizes $\eps$ in $[0,0.4]$ for MNIST and $[0,9/255]$ for CIFAR. For certain networks, we also compute certified accs. using CNN-Cert~\cite{Boopathy2019cnncert} under Fast-Lin type bounds~\cite{weng2018towards} and a variation of CROWN type bounds~\cite{zhang2018crown} with ReLU lower bound of slope zero, which we call CNN-Cert-Zero. Due to the computational cost of these certifiers, we certify on  100 randomly chosen test set points. We use the official code released by~\cite{gowal2019effectiveness} to implement IBP training.

\paragraph{Remark.} In our experiments, we exactly match the parameters used by the authors, with the exception of the number of epochs and batch size used on CIFAR. The authors train networks for $3200$ epochs with a batch size of $1600$, which we find computationally infeasible. Instead, we use the setting of $350$ epochs with a batch size of $50$, which is provided as an alternative in~\cite{gowal2019effectiveness}. This results in some discrepancy with the main results reported by~\cite{gowal2019effectiveness}, but is more consistent with the results reported under the alternative parameter setting in Appendix A of~\cite{gowal2019effectiveness}. In addition, we use IBP a as verifier while~\cite{gowal2019effectiveness} uses exact verifier MILP and consequently gets higher certified accuracy. To avoid other discrepancies, we train IBP networks using code from~\cite{gowal2019effectiveness}, although to ensure a fair comparison, our reported training times are recorded under a consistent implementation of IBP and \singleprop.

\begin{table*}[t]
  \centering
  \caption{Full test set IBP-certified acc. for IBP and \singleprop-trained networks on MNIST and CIFAR-10. Per epoch runtimes are reported, with improvements computed as (IBP runtime/our runtime). For ease of presentation, we use row index to represent the method at that row.}
  \adjustbox{max width = 0.85\textwidth}{
    \begin{tabular}{lrrrrr|r}
    \toprule
    Method & \multicolumn{1}{l}{$\epsilon_{cert}=0$} & \multicolumn{1}{l}{$0.05$} & \multicolumn{1}{l}{$0.10$} & \multicolumn{1}{l}{$0.20$} & \multicolumn{1}{l}{$0.30$} & \multicolumn{1}{l}{Per epoch runtime (s)} \\
    \midrule
    \multicolumn{7}{c}{Small CNN MNIST, 4 layer, $\epsilon_{train}=0.3$} \\
    \midrule
    IBP   & 96.21\% & 95.37\% & 94.35\% & 90.93\% & 84.82\% & \multicolumn{1}{c}{6.3} \\
    \singlepropzero & 94.71\% & 93.70\% & 92.47\% & 88.96\% & 82.93\% & \multicolumn{1}{c}{\textbf{4.0}} \\
    Standard & 99.09\% & 0.00\%  & 0.00\%  & 0.00\%  & 0.00\% & \multicolumn{1}{c}{-}\\
    Adv Training~\cite{madry2017towards} & 99.14\% & 12.26\% & 0.00\%  & 0.00\%  & 0.00\% & \multicolumn{1}{c}{-}\\
    TRADES~\cite{zhang2019theoretically} & 99.09\% & 0.01\%  & 0.00\%  & 0.00\%  & 0.00\% & \multicolumn{1}{c}{-} \\
    \textbf{Improvement:} 
    Row 2 vs. IBP
    & -1.50\% & -1.67\% & -1.88\% & -1.97\% & -1.89\% & \multicolumn{1}{c}{\textbf{\textcolor{blue}{$\times$1.58 faster}}} \\
    \midrule
    \multicolumn{7}{c}{Medium CNN MNIST, 7 layer, $\epsilon_{train}=0.3$} \\
    \midrule
    IBP   & 97.17\% & 96.49\% & 95.59\% & 93.09\% & 88.63\% & \multicolumn{1}{c}{15.0} \\
    \singlepropzero & 97.45\% & 96.55\% & 95.46\% & 92.40\% & 86.05\% & \multicolumn{1}{c}{\textbf{7.1}} \\
    \textbf{Improvement:} Row 2 vs. IBP & \textcolor{blue}{+0.28\%} & \textcolor{blue}{+0.06\%} & -0.13\% & -0.69\% & -2.58\% & \multicolumn{1}{c}{\textbf{\textcolor{blue}{$\times$2.12 faster}}} \\
    \midrule
    \multicolumn{7}{c}{Wide CNN MNIST, 5 layer, $\epsilon_{train}=0.3$} \\
    \midrule
    IBP   & 98.52\% & 98.12\% & 97.36\% & 94.93\% & 89.35\% & \multicolumn{1}{c}{87.7} \\
    \singlepropzero & 97.01\% & 96.27\% & 95.32\% & 92.85\% & 87.59\% & \multicolumn{1}{c}{\textbf{53.1}} \\
    Standard & 99.28\% & 0.00\% & 0.00\% & 0.00\% & 0.00\% & \multicolumn{1}{c}{-} \\
    Adv Training~\cite{madry2017towards} & 99.40\% & 0.00\% & 0.00\% & 0.00\% & 0.00\% & \multicolumn{1}{c}{-} \\
    \textbf{Improvement:} Row 2 vs. IBP & -1.51\% & -1.85\% & -2.04\% & -2.08\% & -1.76\% & \multicolumn{1}{c}{\textbf{\textcolor{blue}{$\times$1.65 faster}}} \\
    \midrule
    Method & \multicolumn{1}{l}{$\epsilon_{cert}=0$} & \multicolumn{1}{l}{$2/255$} & \multicolumn{1}{l}{$5/255$} & \multicolumn{1}{l}{$7/255$} & \multicolumn{1}{l}{$8/255$} & \multicolumn{1}{l}{Per epoch runtime (s)} \\
    \midrule
    \multicolumn{7}{c}{Small CNN CIFAR, 4 layer, $\epsilon_{train}=8/255$} \\
    \midrule
    IBP & 36.46\% & 33.70\% & 29.64\% & 27.14\% & 25.99\% & \multicolumn{1}{c}{14.8} \\
    \singlepropfastlin & 37.79\% & 34.39\% & 29.01\% & 26.04\% & 24.51\% & \multicolumn{1}{c}{\textbf{7.2}} \\
    \textbf{Improvement:} Row 2 vs. IBP & \textcolor{blue}{+1.33\%} & \textcolor{blue}{+0.69\%} & -0.63\% & -1.10\% & -1.48\% & \multicolumn{1}{c}{\textbf{\textcolor{blue}{$\times$2.07 faster}}} \\
    \midrule
    \multicolumn{7}{c}{Large CNN CIFAR, 7 layer, $\epsilon_{train}=8/255$} \\
    \midrule
    IBP   & 46.80\% & 41.15\% & 33.16\% & 28.04\% & 25.68\% & \multicolumn{1}{c}{43.7} \\
    \singlepropfastlin & 44.36\% & 37.79\% & 29.51\% & 24.28\% & 21.94\% & \multicolumn{1}{c}{\textbf{21.6}} \\
    \textbf{Improvement:} Row 2 vs. IBP & -2.44\% & -3.36\% & -3.65\% & -3.76\% & -3.74\% & \multicolumn{1}{c}{\textbf{\textcolor{blue}{$\times$2.02 faster}}} \\
    \bottomrule
    \end{tabular}%
  \label{tab:main}%
  }
\end{table*}%

\subsection{Results}
\paragraph{Results on MNIST and CIFAR.}
In Table~\ref{tab:main} we find that on MNIST, \singlepropzero achieves similar performance to IBP on both clean acc. and robust acc. over all $\eps$. Specifically, on the Small model, \singlepropzero achieves a clean acc. of $94.71\%$ and a certified acc. of $82.93 \%$ at $\eps = 0.3$ compared to a clean acc. of $96.21\%$ and a certified acc. of $84.82 \%$ for IBP. We observe similar results under an average of 5 trials (see App.~\ref{app:tables} Table~\ref{tab:complementary}), with a maximum certified accuracy standard deviation of $2.36\%$. On the Medium model, \singlepropzero achieves a clean acc. of $97.45\%$ and a certified acc. of $86.05 \%$ at $\eps = 0.3$ compared to a clean acc. of $97.17\%$ and a certified acc. of $88.63 \%$ for IBP\footnote{We note that~\cite{gowal2019effectiveness} report higher certified accs. ($91.95 \%$ on MNIST and $32.04 \%$ on CIFAR). On CIFAR, the authors train for $3200$ epochs, which we find computationally infeasible. With $350$ epochs, the authors report a certified acc. of $28.30 \%$ which is more comparable with our results.}. On the Wide model, \singlepropzero achieves a clean acc. of $97.01\%$ and a certified acc. of $87.59 \%$ at $\eps = 0.3$ compared to a clean acc. of $98.52\%$ and a certified acc. of $89.35 \%$ for IBP.

For the CIFAR Small model, \singlepropfastlin achieves similar clean acc. and robust acc. over all $\eps$ under multiple trials (see App.~\ref{app:tables} Table~\ref{tab:complementary}). Specifically, on the Small model, \singlepropfastlin achieves a clean acc. range of $[36.97 \%, 37.79 \%]$ and a certified acc. range of $[23.94 \%, 24.71 \%]$ at $\eps = 8/255$ compared to a clean acc. range of $[36.46 \%, 39.05 \%]$ and a certified acc. of $[25.99 \%, 26.57 \%]$ for IBP. Interestingly, we observe that certified accs. are slightly lower on the Large model, but clean accs. are much larger with $\singleprop$ achieving a clean acc. of $44.36 \%$ and a certified acc. of $21.94 \%$ at $\eps = 8/255$ compared to a clean acc. of $46.80 \%$ and a certified acc. of $25.68 \%$ for IBP.

\paragraph{Runtime Improvement.}
As seen in the last column of Table~\ref{tab:main}, \singleprop consistently trains in less time than IBP, with between $1.54\times$ and $2.24\times$ speedup (measured as IBP runtime/our runtime). This performance is faster than expected by the number of forward propagations: since IBP uses three forward propagations compared to two for \singleprop, a speedup of $1.5\times$ would be expected. Slower methods such as IBP or CROWN-IBP can achieve higher certified accuracies at the cost of computational efficiency, corresponding to different points on a trade-off curve between training time and certification level (see App.~\ref{app:tables} Table~\ref{tab:tradeoff}). Thus, \singleprop may be preferable to other methods when considering computation and memory costs.

\paragraph{\singleprop approximation error.} \label{sec: approx_error}
We evaluate the quality of the approximation in Eq.~\eqref{eqn:v_def} which is used to derive \singleprop as an approximation of linear bounding certifiers. We compute two metrics: Metric 1 = $\sum_{i,j} |\z^i_{nom,j} - (\z^i_{U,j} + \z^i_{L,j})/2|$, metric 2 = $\sum_{i,j} |\z^i_{nom,j} - (\z^i_{U,j} + \z^i_{L,j})/2|/(\z^i_{U,j} - \z^i_{L,j})$. On Small CNN MNIST models, the (mean, std) of metric 1 is IBP: (8.83E-07, 1.57E-07), \singlepropzero: (4.08E-07, 3.80E-08), and on metric 2 is IBP: (0.018, 0.016), \singlepropzero: (0.003, 0.004). This shows that $\singlepropzero$ indeed closely approximates upper and lower bounds on a linear bounding certifier on models trained with $\singlepropzero$ or IBP.

\paragraph{Combining model certifications greatly improves certified accuracies.}
As observed in App.~\ref{app:tables} Table~\ref{tab:complementary}, combining the points certified under IBP and \singleprop models individually increases certified accs. by $2$-$4\%$ at $\eps = 0.3$ and by $6$-$9\%$ at $\eps = 8/255$ on CIFAR. In other words, points uncertifiable by an IBP network are certifiable under a \singleprop network and vice versa. This demonstrates that IBP and \singlepropfastlin networks complement each other in their certifications. We also evaluate the complementarity of IBP and \singleprop certifications in the set of points correctly classified by both methods ($\{ C \}$ in Table~\ref{tab:complementary}), and find that nearly all points in this set are certifiable under at least one model. We note that bound complementarity is greater between IBP and \singlepropfastlin (compared to \singlepropzero). This may be because while \singlepropzero approximates IBP (see Sec~\ref{sec:regularizer}), \singlepropfastlin approximates Fast-Lin, encouraging certifiability under Fast-Lin, which verifies different points than IBP. Therefore, \singlepropfastlin could induce robustness in points that are difficult to certify under IBP-like training, leading to certifiability even under IBP verification.

\paragraph{Evaluating under other certifiers.}
In App.~\ref{app:tables} Table~\ref{tab:100} we find that IBP and CNN-Cert-Zero certified accs. are similar for both IBP and \singlepropzero trained models. Because CNN-Cert-Zero bounds are stronger, but similar to IBP (see Sec~\ref{sec:regularizer}),
\singlepropzero can achieve high acc. even under IBP certification, and IBP also performs similarly on CNN-Cert-Zero as IBP certification. We also observe that under Fast-Lin, IBP networks perform worse than \singlepropfastlin. In particular, \singlepropfastlin maintains similar accs. under Fast-Lin as IBP, validating that \singlepropfastlin is indeed adapted for Fast-Lin. Finally, including points certifiable by either verifier (Fast-Lin+IBP) results in similar accs. as only using IBP, justifying using IBP as our primary verifier. In App.~\ref{app:tables} Table~\ref{tab:100vs200}, we also evaluate certified accuracies on 200 points for selected networks and find similar results to 100 points.

\paragraph{Evaluating Adaptive Hyperparameter Selection (AHS).}
We evaluate the effect of adaptive hyperparameter selection on both IBP and \singleprop training by comparing AHS to the tuned piecewise linear schedule for $\lambda$ used in~\cite{gowal2019effectiveness}. In App.~\ref{app:tables} Table~\ref{tab:ada}, we show certified accs. under both schemes, reporting the best results under a grid search for learning rates. The piecewise linear schedule is denoted Linear. As illustrated, with the exception of the IBP models trained on Small CNN MNIST, adaptive hyperparameter selection increases certified accs. at $\eps=0.3$ for MNIST and $\eps=8/255$ for CIFAR. Moreover, for \singleprop models, certified accs. increase for most $\eps$ by up to about $1 \%$, while decreasing at most by $0.05\%$. This indicates that AHS can achieve high robust accs. while avoiding tuning over the large space of hyperparameter schedules.

\mycomment{
\PY{Some comments:\\
1. Table captions should be put above table, not below
\\
2. We argue a lot on computation time saving, but I don't see any run-time benchmarks. The run time analysis should be the main different here
}}

\section{Conclusion}
In this paper, we propose an efficient certified training framework, \singleprop, that requires only one additional forward propagation through a network. We have conducted a comprehensive comparison on \singleprop and current SOTA most efficient certified training framework, IBP, in terms of the training schedules, verifiers and complementary verification results. We found that we achieve comparable certified accuracies to state-of-the-art certified defenses while being $1.5$-$2\times$ faster to train and requires $1.5 \times$ less memory usage. To our best knowledge, \textbf{\singleprop} is the fastest and most efficient among SOTA certified training algorithms.  

\mycomment{
\Lily{needs to be rewritten.}
In this paper, we propose a unified framework of certified defense as regularization. We introduce a new efficient regularizer under this framework, and extend it to a $\ell_0$ threat model and model ensembles. Through comparative experiments on several model architectures, we demonstrate that our method outperforms state-of-the-art certification based training methods in terms of training time and certified accuracy.}

\section*{Ethical Impact}
This work presents a method of efficiently training networks that are verifiable against adversarial attacks. Verifiably robust networks may be important in safety-critical scenarios such as self-driving cars or medical diagnosis where not only must models be robust to adversarial attack, but robustness must be verifiable to establish trust. Our method could help expand access to verifiable models in cases where training verifiable models is difficult due to computational constraints. This includes situations like image classifiers on real world images, where training high-accuracy, verifiable models may be out of reach due to computational cost. Our work presents potential benefits for users and creators of safety-critical models.

At the same time, it is possible that reliance on verification may provide a false sense of security to users who are not familiar with verification techniques. For instance, users may interpret verified model outputs as applying to a broader range of adversarial perturbations than the specific perturbation norm that is verified. As such, this work might potentially harm users less familiar with adversarial robustness and verification. We believe properly communicating the type and level of robustness implied by verification will be important in reducing the risks of our work.

\mycomment{\SL{[Impact statement is required and can be put at page 9.]}}

\bibliography{ref}

\begin{thebibliography}{38}
\providecommand{\natexlab}[1]{#1}
\providecommand{\url}[1]{\texttt{#1}}
\providecommand{\urlprefix}{URL }
\expandafter\ifx\csname urlstyle\endcsname\relax
  \providecommand{\doi}[1]{doi:\discretionary{}{}{}#1}\else
  \providecommand{\doi}{doi:\discretionary{}{}{}\begingroup
  \urlstyle{rm}\Url}\fi

\bibitem[{Abadi et~al.(2016)Abadi, Barham, Chen, Chen, Davis, Dean, Devin,
  Ghemawat, Irving, Isard et~al.}]{abadi2016tensorflow}
Abadi, M.; Barham, P.; Chen, J.; Chen, Z.; Davis, A.; Dean, J.; Devin, M.;
  Ghemawat, S.; Irving, G.; Isard, M.; et~al. 2016.
\newblock TensorFlow: A System for Large-Scale Machine Learning.

\bibitem[{Athalye, Carlini, and Wagner(2018)}]{athalye2018obfuscated}
Athalye, A.; Carlini, N.; and Wagner, D. 2018.
\newblock Obfuscated Gradients Give a False Sense of Security: Circumventing
  Defenses to Adversarial Examples.
\newblock \emph{ICML} .

\bibitem[{Athalye and Sutskever(2017)}]{athalye2017synthesizing}
Athalye, A.; and Sutskever, I. 2017.
\newblock Synthesizing robust adversarial examples.
\newblock \emph{arXiv preprint arXiv:1707.07397} .

\bibitem[{Boopathy et~al.(2019)Boopathy, Weng, Chen, Liu, and
  Daniel}]{Boopathy2019cnncert}
Boopathy, A.; Weng, T.-W.; Chen, P.-Y.; Liu, S.; and Daniel, L. 2019.
\newblock CNN-Cert: An Efficient Framework for Certifying Robustness of
  Convolutional Neural Networks.
\newblock In \emph{AAAI}.

\bibitem[{Carlini et~al.(2017)Carlini, Katz, Barrett, and
  Dill}]{carlini2017ground}
Carlini, N.; Katz, G.; Barrett, C.; and Dill, D.~L. 2017.
\newblock Provably Minimally-Distorted Adversarial Examples.
\newblock \emph{arXiv preprint arXiv:1709.10207} .

\bibitem[{Carlini and Wagner(2017)}]{carlini2017towards}
Carlini, N.; and Wagner, D. 2017.
\newblock Towards evaluating the robustness of neural networks.
\newblock In \emph{IEEE Symposium on Security and Privacy (SP)}, 39--57.

\bibitem[{Chen et~al.(2018)Chen, Sharma, Zhang, Yi, and Hsieh}]{chen2017ead}
Chen, P.-Y.; Sharma, Y.; Zhang, H.; Yi, J.; and Hsieh, C.-J. 2018.
\newblock EAD: elastic-net attacks to deep neural networks via adversarial
  examples.
\newblock \emph{AAAI} .

\bibitem[{Cheng, N{\"u}hrenberg, and Ruess(2017)}]{cheng2017maximum}
Cheng, C.-H.; N{\"u}hrenberg, G.; and Ruess, H. 2017.
\newblock Maximum resilience of artificial neural networks.
\newblock In \emph{International Symposium on Automated Technology for
  Verification and Analysis}, 251--268. Springer.

\bibitem[{Dvijotham et~al.(2018{\natexlab{a}})Dvijotham, Gowal, Stanforth,
  Arandjelovic, O'Donoghue, Uesato, and Kohli}]{dvijotham2018training}
Dvijotham, K.; Gowal, S.; Stanforth, R.; Arandjelovic, R.; O'Donoghue, B.;
  Uesato, J.; and Kohli, P. 2018{\natexlab{a}}.
\newblock Training verified learners with learned verifiers.
\newblock \emph{arXiv preprint arXiv:1805.10265} .

\bibitem[{Dvijotham et~al.(2018{\natexlab{b}})Dvijotham, Stanforth, Gowal,
  Mann, and Kohli}]{dvijotham2018dual}
Dvijotham, K.; Stanforth, R.; Gowal, S.; Mann, T.; and Kohli, P.
  2018{\natexlab{b}}.
\newblock A dual approach to scalable verification of deep networks.
\newblock \emph{UAI} .

\bibitem[{Gehr et~al.(2018)Gehr, Mirman, Drachsler-Cohen, Tsankov, Chaudhuri,
  and Vechev}]{Gehr2018AI2}
Gehr, T.; Mirman, M.; Drachsler-Cohen, D.; Tsankov, P.; Chaudhuri, S.; and
  Vechev, M. 2018.
\newblock AI2: Safety and Robustness Certification of Neural Networks with
  Abstract Interpretation.
\newblock In \emph{IEEE Symposium on Security and Privacy (SP)}, volume~00,
  948--963.

\bibitem[{Goodfellow, Shlens, and Szegedy(2015)}]{goodfellow2014explaining}
Goodfellow, I.~J.; Shlens, J.; and Szegedy, C. 2015.
\newblock Explaining and harnessing adversarial examples.
\newblock \emph{ICLR} .

\bibitem[{Gowal et~al.(2019)Gowal, Dvijotham, Stanforth, Bunel, Qin, Uesato,
  Arandjelovic, Mann, and Kohli}]{gowal2019effectiveness}
Gowal, S.; Dvijotham, K.; Stanforth, R.; Bunel, R.; Qin, C.; Uesato, J.;
  Arandjelovic, R.; Mann, T.~A.; and Kohli, P. 2019.
\newblock Scalable Verified Training for Provably Robust Image Classification.
\newblock \emph{ICCV} .

\bibitem[{Hein and Andriushchenko(2017)}]{hein2017formal}
Hein, M.; and Andriushchenko, M. 2017.
\newblock Formal guarantees on the robustness of a classifier against
  adversarial manipulation.
\newblock In \emph{NIPS}.

\bibitem[{Katz et~al.(2017)Katz, Barrett, Dill, Julian, and
  Kochenderfer}]{katz2017reluplex}
Katz, G.; Barrett, C.; Dill, D.~L.; Julian, K.; and Kochenderfer, M.~J. 2017.
\newblock Reluplex: An efficient SMT solver for verifying deep neural networks.
\newblock In \emph{International Conference on Computer Aided Verification},
  97--117. Springer.

\bibitem[{Kingma and Ba(2014)}]{kingma2014adam}
Kingma, D.; and Ba, J. 2014.
\newblock Adam: A method for stochastic optimization.
\newblock \emph{arXiv preprint arXiv:1412.6980} .

\bibitem[{Ko et~al.(2019)Ko, Lyu, Weng, Daniel, Wong, and Lin}]{ko2019popqorn}
Ko, C.-Y.; Lyu, Z.; Weng, T.-W.; Daniel, L.; Wong, N.; and Lin, D. 2019.
\newblock POPQORN: Quantifying robustness of recurrent neural networks.
\newblock \emph{arXiv preprint arXiv:1905.07387} .

\bibitem[{Kolter and Wong(2018)}]{kolter2017provable}
Kolter, J.~Z.; and Wong, E. 2018.
\newblock Provable defenses against adversarial examples via the convex outer
  adversarial polytope.
\newblock \emph{ICML} .

\bibitem[{Madry et~al.(2018)Madry, Makelov, Schmidt, Tsipras, and
  Vladu}]{madry2017towards}
Madry, A.; Makelov, A.; Schmidt, L.; Tsipras, D.; and Vladu, A. 2018.
\newblock Towards Deep Learning Models Resistant to Adversarial Attacks.
\newblock \emph{ICLR} .

\bibitem[{Mirman, Gehr, and Vechev(2018)}]{mirman2018differentiable}
Mirman, M.; Gehr, T.; and Vechev, M. 2018.
\newblock Differentiable abstract interpretation for provably robust neural
  networks.
\newblock In \emph{International Conference on Machine Learning}, 3575--3583.

\bibitem[{Papernot et~al.(2016)Papernot, McDaniel, Jha, Fredrikson, Celik, and
  Swami}]{papernot2016limitations}
Papernot, N.; McDaniel, P.; Jha, S.; Fredrikson, M.; Celik, Z.~B.; and Swami,
  A. 2016.
\newblock The limitations of deep learning in adversarial settings.
\newblock In \emph{IEEE European Symposium on Security and Privacy (EuroS\&P)},
  372--387.

\bibitem[{Peck et~al.(2017)Peck, Roels, Goossens, and Saeys}]{peck2017lower}
Peck, J.; Roels, J.; Goossens, B.; and Saeys, Y. 2017.
\newblock Lower bounds on the robustness to adversarial perturbations.
\newblock In \emph{NIPS}.

\bibitem[{Raghunathan, Steinhardt, and Liang(2018)}]{raghunathan2018certified}
Raghunathan, A.; Steinhardt, J.; and Liang, P. 2018.
\newblock Certified defenses against adversarial examples.
\newblock \emph{ICLR} .

\bibitem[{{Salman} et~al.(2019){Salman}, {Yang}, {Zhang}, {Hsieh}, and
  {Zhang}}]{salman2019convex}
{Salman}, H.; {Yang}, G.; {Zhang}, H.; {Hsieh}, C.-J.; and {Zhang}, P. 2019.
\newblock {A Convex Relaxation Barrier to Tight Robustness Verification of
  Neural Networks}.
\newblock \emph{arXiv preprint arXiv:1902.08722} .

\bibitem[{Singh et~al.(2018)Singh, Gehr, Mirman, P\"{u}schel, and
  Vechev}]{singh2018fast}
Singh, G.; Gehr, T.; Mirman, M.; P\"{u}schel, M.; and Vechev, M. 2018.
\newblock Fast and Effective Robustness Certification.
\newblock In \emph{NeurIPS}.

\bibitem[{Singh et~al.(2019)Singh, Gehr, Mirman, P\"{u}schel, and
  Vechev}]{singh2019abstract}
Singh, G.; Gehr, T.; Mirman, M.; P\"{u}schel, M.; and Vechev, M. 2019.
\newblock An Abstract Domain for Certifying Neural Networks.
\newblock In \emph{POPL}.

\bibitem[{Sinha, Namkoong, and Duchi(2018)}]{sinha2017certifiable}
Sinha, A.; Namkoong, H.; and Duchi, J. 2018.
\newblock Certifiable Distributional Robustness with Principled Adversarial
  Training.
\newblock \emph{ICLR} .

\bibitem[{Su, Vargas, and Sakurai(2019)}]{su2019one}
Su, J.; Vargas, D.~V.; and Sakurai, K. 2019.
\newblock One pixel attack for fooling deep neural networks.
\newblock \emph{IEEE Transactions on Evolutionary Computation} .

\bibitem[{Szegedy et~al.(2013)Szegedy, Zaremba, Sutskever, Bruna, Erhan,
  Goodfellow, and Fergus}]{szegedy2013intriguing}
Szegedy, C.; Zaremba, W.; Sutskever, I.; Bruna, J.; Erhan, D.; Goodfellow, I.;
  and Fergus, R. 2013.
\newblock Intriguing properties of neural networks.
\newblock \emph{arXiv preprint arXiv:1312.6199} .

\bibitem[{Tjeng and Tedrake(2019)}]{tjeng2019evaluating}
Tjeng, V.; and Tedrake, R. 2019.
\newblock Verifying Neural Networks with Mixed Integer Programming.
\newblock In \emph{ICLR}.

\bibitem[{Wang et~al.(2018)Wang, Pei, Whitehouse, Yang, and
  Jana}]{wang2018efficient}
Wang, S.; Pei, K.; Whitehouse, J.; Yang, J.; and Jana, S. 2018.
\newblock Efficient Formal Safety Analysis of Neural Networks.
\newblock In \emph{NeurIPS}.

\bibitem[{Weng et~al.(2018)Weng, Zhang, Chen, Song, Hsieh, Boning, Dhillon, and
  Daniel}]{weng2018towards}
Weng, T.-W.; Zhang, H.; Chen, H.; Song, Z.; Hsieh, C.-J.; Boning, D.; Dhillon,
  I.~S.; and Daniel, L. 2018.
\newblock Towards Fast Computation of Certified Robustness for ReLU Networks.
\newblock \emph{ICML} .

\bibitem[{Wong et~al.(2018)Wong, Schmidt, Metzen, and Kolter}]{wong2018scaling}
Wong, E.; Schmidt, F.; Metzen, J.~H.; and Kolter, J.~Z. 2018.
\newblock Scaling provable adversarial defenses.
\newblock \emph{arXiv preprint arXiv:1805.12514} .

\bibitem[{Xiao et~al.(2019)Xiao, Tjeng, Shafiullah, and
  Madry}]{xiao2019training}
Xiao, K.~Y.; Tjeng, V.; Shafiullah, N.~M.; and Madry, A. 2019.
\newblock Training for Faster Adversarial Robustness Verification via Inducing
  ReLU Stability.
\newblock In \emph{ICLR}.

\bibitem[{Xu et~al.(2018)Xu, Liu, Zhao, Chen, Zhang, Erdogmus, Wang, and
  Lin}]{xu2018structured}
Xu, K.; Liu, S.; Zhao, P.; Chen, P.-Y.; Zhang, H.; Erdogmus, D.; Wang, Y.; and
  Lin, X. 2018.
\newblock Structured adversarial attack: Towards general implementation and
  better interpretability.
\newblock \emph{arXiv preprint arXiv:1808.01664} .

\bibitem[{Zhang et~al.(2020)Zhang, Chen, Xiao, Li, Boning, and
  Hsieh}]{zhang2020crownibp}
Zhang, H.; Chen, H.; Xiao, C.; Li, B.; Boning, D.~S.; and Hsieh, C. 2020.
\newblock Towards Stable and Efficient Training of Verifiably Robust Neural
  Networks.
\newblock \emph{ICLR} .

\bibitem[{Zhang et~al.(2018)Zhang, Weng, Chen, Hsieh, and
  Daniel}]{zhang2018crown}
Zhang, H.; Weng, T.-W.; Chen, P.-Y.; Hsieh, C.-J.; and Daniel, L. 2018.
\newblock Efficient Neural Network Robustness Certification with General
  Activation Functions.
\newblock In \emph{NIPS}.

\bibitem[{{Zhang} et~al.(2019){Zhang}, {Yu}, {Jiao}, {Xing}, {El Ghaoui}, and
  {Jordan}}]{zhang2019theoretically}
{Zhang}, H.; {Yu}, Y.; {Jiao}, J.; {Xing}, E.~P.; {El Ghaoui}, L.; and
  {Jordan}, M.~I. 2019.
\newblock {Theoretically Principled Trade-off between Robustness and Accuracy}.
\newblock In \emph{ICML}.

\end{thebibliography}

\clearpage

\newpage

\onecolumn
\section*{Appendix}
\appendix
\section{Table of Notation} \label{app:notation}
\begin{table*}[ht!]
  \centering
  \caption{Table of Notation}
  \adjustbox{max width = 0.9\textwidth}{
    \begin{tabular}{clcl}
    \toprule
    \textbf{Notation} & \textbf{Definition} & \textbf{Notation} & \textbf{Definition}\\
    \hline
    \addlinespace[0.1em]
    $\xnom$ & unperturbed input & $\z_L^i$ & lower bound of layer $i$\\
    \addlinespace[0.1em]
    $\delta$ & input perturbation & $\z_U^i$ & upper bound of layer $i$\\
    \addlinespace[0.1em]
    $\eps$ & maximum $\ell_p$ perturbation size $||\delta||_p$ & $\bu^i,\bl^i$ & pre-activation interval bounds\\
    \addlinespace[0.1em]
    $\x$ & perturbed input & $\bv^i$ & layerwise uncertainty quantifier\\
    \addlinespace[0.1em]
    $f$ & neural network classifier & $\alpha_U^i, \alpha_L^i, \beta_U^i, \beta_L^i$ & linear bounding parameters\\
    \addlinespace[0.1em]
    $\sigma$ & neural network activation & $\A_U, \A_L ,\bias_L,\bias_U$ & linear network bounds\\
    \addlinespace[0.1em]
    $n$ & number of network layers & $\loss : (\z^n ,\mathbf{y}) \rightarrow \real$ & training loss function\\
    \addlinespace[0.1em]
    $\theta$ & model parameters  & $\loss_{val}: (\z^n ,\mathbf{y}) \rightarrow \real$ & validation loss function\\
    \addlinespace[0.1em]
    $\W^i,\bias^i$ & weights, biases of layer $i$  & $\reg: (\cdot, \cdot, \mathbf{y}) \rightarrow \real$ & regularizer\\
    \addlinespace[0.1em]
    $\z^i$ & input of layer $i$ & $\lam$ & regularizer coefficient\\
    \addlinespace[0.1em]
    \bottomrule
    \end{tabular}%
  }
  \label{tab:Notation_updated}%
\end{table*}%

\section{\singleprop Algorithm} \label{app:algo}

\label{algo}
\begin{algorithm}[h!]
\SetAlgoLined
\KwData{Training data $\mathcal{D}$, Randomly initialized weights and biases $\W^i,\bias^i,i=1:N$, Number of epochs $E$, Perturbation size schedule $\eps(t)$, Regularization weight schedule $\lam(t)$, Adaptive hyperparameter flag \texttt{ada}}
\KwResult{Trained weights and biases $\W^i,\bias^i$}
If {\texttt{ada}} initialize $\lam = 0$

Initialize $t = 0$

 \For{epoch = 1:E}{
    \For{$\xnom$ in $\mathcal{D}$}{
        $\z^0 = \xnom,\bv^0 = \eps(t) \mathbf{1}$\;
        
        \For{i = 1:N}{
            Propagate $\z^{i+1}$ with Equation~\eqref{eqn:z_def}

            Propagate $\bv^{i+1}$ with
            Equation~\eqref{eqn:v_def}
        }
        Construct regularization term $\reg$ according to \eqref{eq:single_margin_regularizer_v2}
        
        If not {\texttt{ada}} $\lam = \lam(t)$
        
        Compute gradients of $\loss+\lam \reg$ w.r.t. $\W^i,\bias^i$
        
        Update $\W^i,\bias^i$ via gradient descent
        
        $t = t + 1$
    }
    If {\texttt{ada}} $\lambda = \frac{\gamma \loss_{val,\theta}(\z^n, \mathbf{y})}{(1+\gamma)\loss_{val,\theta}(\z^n, \mathbf{y}) + \reg_{val,\theta}(\cdot,\cdot, \mathbf{y})}$
 }
 \caption{\singleprop Training Algorithm}
 \label{algo:singlemargin_algo}
\end{algorithm}

\section{Implementation Details} \label{app:impl}
Training methods are implemented in Python with Tensorflow ~\citep{abadi2016tensorflow} and training is conducted on a NVIDIA Tesla V100 GPU. The Small CNN architecture uses convolution layers of 16 and 32 filters followed by a 100 unit fully connected layer. The Medium CNN architecture uses convolutional layers of 16, 16, 32 and 32 filters followed by two 512 unit fully connected layers. The Large CNN architecture uses convolutional layers of 64, 64, 64, 128 and 128 filters followed by a 200 unit fully connected layer. The Wide CNN architecture uses convolutional layers of 128, 256 and 512 layers followed by a 1024 unit fully connected layer. Following~\cite{gowal2019effectiveness}, networks are trained with Adam~\cite{kingma2014adam} with the learning rate decayed by a factor of 10 twice at $15000$ and $25000$ training steps on MNIST and $60000$ and $90000$ training steps on CIFAR. A grid search is conducted over the choice of initial learning rate with range $[0.0002, 0.01]$ and whether to use a piece-wise linear or adaptive schedule for $\lambda$. We report results for networks achieving the highest robust test accuracy at $\eps = 0.3$ for MNIST and $\eps =  8/255$ for CIFAR.

\section{Additional Tables}\label{app:tables}

\begin{table}[htbp]
  \centering
  \caption{Comparison of SingleProp, IBP and CROWN-IBP for Small CNN CIFAR on certified accuracy and additional computational complexity relative to standard training}
    \begin{tabular}{rrl}
    \toprule
    Method & {Certified accuracy at $\epsilon=$8/255} & {Computational Complexity (cost)} \\
    \midrule
    \singlepropfastlin & 24.51\% & 1x \\
    IBP   & 25.99\% & 2x \\
    CROWN-IBP & 29.24\% & 26x \\
    \bottomrule
    \end{tabular}%
  \label{tab:tradeoff}%
\end{table}%

\begin{table*}[htbp]
  \centering
  \caption{IBP certified test set accuracies on networks trained with IBP and \singleprop on MNIST and CIFAR-10. Both methods are trained under a piecewise-linear schedule for regularization parameter $\lambda$ (denoted Linear) and an adaptive hyperparameter selection scheme based valdiation set performance (denoted AHS). Runtimes are reported per training epoch.}
  \adjustbox{max width = \textwidth}{
        \begin{tabular}{lrrrrrrrrr|c}
    \toprule
    Method & \multicolumn{1}{l}{$\epsilon_{cert}=0$} & \multicolumn{1}{l}{$0.01$} & \multicolumn{1}{l}{$0.03$} & \multicolumn{1}{l}{$0.05$} & \multicolumn{1}{l}{$0.07$} & \multicolumn{1}{l}{$0.10$} & \multicolumn{1}{l}{$0.20$} & \multicolumn{1}{l}{$0.30$} & \multicolumn{1}{l}{$0.40$} & \multicolumn{1}{l}{Per epoch runtime (s)} \\
    \midrule
    \multicolumn{11}{c}{Small CNN MNIST, 4 layer, $\epsilon_{train}=0.3$} \\
    \midrule
    IBP, Linear & 97.18\% & 97.03\% & 96.73\% & 96.39\% & 95.94\% & 95.27\% & 92.04\% & 85.16\% & 0.00\% & 5.9 \\
    IBP, AHS & 96.21\% & 96.10\% & 95.83\% & 95.37\% & 94.96\% & 94.35\% & 90.93\% & 84.82\% & 0.00\% & 6.3 \\
    \singlepropzero, Linear & 94.73\% & 94.51\% & 94.05\% & 93.48\% & 93.01\% & 92.02\% & 87.91\% & 81.85\% & 0.00\% & \textbf{3.7} \\
    \singlepropzero, AHS & 94.71\% & 94.52\% & 94.15\% & 93.70\% & 93.29\% & 92.47\% & 88.96\% & 82.93\% & 0.00\% & \textbf{4.0} \\
    \textbf{Improv.}: Row 3 vs. IBP, Linear & -2.45\% & -2.52\% & -2.68\% & -2.91\% & -2.93\% & -3.25\% & -4.13\% & -3.31\% & 0.00\% & \textbf{\textcolor{blue}{$\times$1.59 faster}}\\
    \textbf{Improv.}: Row 4 vs. IBP, AHS & -1.50\% & -1.58\% & -1.68\% & -1.67\% & -1.67\% & -1.88\% & -1.97\% & -1.89\% & 0.00\% & \textbf{\textcolor{blue}{$\times$1.58 faster}}\\
    \midrule
    \multicolumn{11}{c}{Medium CNN MNIST, 7 layer, $\epsilon_{train}=0.3$} \\
    \midrule
    IBP, Linear & 96.95\% & 96.79\% & 96.41\% & 96.01\% & 95.65\% & 95.07\% & 92.25\% & 87.11\% & 0.00\% & 8.8 \\
    IBP, AHS & 97.17\% & 97.04\% & 96.78\% & 96.49\% & 96.13\% & 95.59\% & 93.09\% & 88.63\% & 0.00\% & 8.9 \\
    \singlepropzero, Linear & 97.38\% & 97.25\% & 96.87\% & 96.45\% & 96.14\% & 95.35\% & 92.02\% & 85.34\% & 0.01\% & \textbf{5.5} \\
    \singlepropzero, AHS & 97.45\% & 97.22\% & 96.91\% & 96.55\% & 96.09\% & 95.46\% & 92.40\% & 86.05\% & 0.00\% & \textbf{5.8} \\
    \textbf{Improv.}: Row 3 vs. IBP, Linear & \textcolor{blue}{+0.43\%} & \textcolor{blue}{+0.46\%} & \textcolor{blue}{+0.46\%} & \textcolor{blue}{+0.44\%} & \textcolor{blue}{+0.49\%} & \textcolor{blue}{+0.28\%} & -0.23\% & -1.77\% & \textcolor{blue}{+0.01\%} & \textbf{\textcolor{blue}{$\times$1.59 faster}}\\
    \textbf{Improv.}: Row 4 vs. IBP, AHS & 0.28\% & 0.18\% & 0.13\% & 0.06\% & -0.04\% & -0.13\% & -0.69\% & -2.58\% & 0.00\% & \textbf{\textcolor{blue}{$\times$1.54 faster}}\\
    \midrule
    Method & \multicolumn{1}{l}{$\epsilon_{cert}=0$} & \multicolumn{1}{l}{$0.5/255$} & \multicolumn{1}{l}{$1/255$} & \multicolumn{1}{l}{$2/255$} & \multicolumn{1}{l}{$3/255$} & \multicolumn{1}{l}{$5/255$} & \multicolumn{1}{l}{$7/255$} & \multicolumn{1}{l}{$8/255$} & \multicolumn{1}{l}{$9/255$} & \multicolumn{1}{l}{Per epoch runtime (s)} \\
    \midrule
    \multicolumn{11}{c}{Small CNN CIFAR, 4 layer, $\epsilon_{train}=8/255$} \\
    \midrule
    IBP, Linear & 42.82\% & 41.53\% & 40.32\% & 38.08\% & 35.88\% & 31.26\% & 26.88\% & 24.91\% & 23.08\% & 14.6 \\
    IBP, AHS & 38.05\% & 37.30\% & 36.51\% & 35.08\% & 33.84\% & 30.82\% & 27.97\% & 26.57\% & 25.11\% & 15.0 \\
    \singlepropfastlin, Linear & 35.98\% & 35.18\% & 34.34\% & 32.73\% & 30.95\% & 27.78\% & 24.91\% & 23.63\% & 22.21\% & \textbf{6.5} \\
    \singlepropfastlin, AHS & 36.97\% & 36.14\% & 35.24\% & 33.80\% & 32.09\% & 28.55\% & 25.25\% & 23.94\% & 22.69\% & \textbf{7.1} \\
    \textbf{Improv.}: Row 3 vs. IBP, Linear & -6.84\% & -6.35\% & -5.98\% & -5.35\% & -4.93\% & -3.48\% & -1.97\% & -1.28\% & -0.87\% & \textbf{\textcolor{blue}{$\times$2.24 faster}}\\
    \textbf{Improv.}: Row 4 vs. IBP, AHS & -1.08\% & -1.16\% & -1.27\% & -1.28\% & -1.75\% & -2.27\% & -2.72\% & -2.63\% & -2.42\% & \textbf{\textcolor{blue}{$\times$2.12 faster}}\\
    \bottomrule
    \end{tabular}%
    }
  \label{tab:ada}%
\end{table*}%

\begin{table*}[htbp]
  \centering
  \caption{Certified accuracies on networks trained with IBP and \singleprop on MNIST and CIFAR-10. Certification is performed with IBP,  Fast-Lin and CNN-Cert-Zero certification as well as Fast-Lin+IBP which includes points certifiable by either method. Accuracies are reported on 100 random test set points. \textcolor[rgb]{ 1,  0,  0}{Red} numbers indicate significantly worse performance in comparison to other verifiers.}
  \adjustbox{max width = \textwidth}{
    \begin{tabular}{lrrrrrrrrr}
    \toprule
    Method (100 points) & \multicolumn{1}{l}{$\epsilon_{cert}=0$} & \multicolumn{1}{l}{$0.01$} & \multicolumn{1}{l}{$0.03$} & \multicolumn{1}{l}{$0.05$} & \multicolumn{1}{l}{$0.07$} & \multicolumn{1}{l}{$0.10$} & \multicolumn{1}{l}{$0.20$} & \multicolumn{1}{l}{$0.30$} & \multicolumn{1}{l}{$0.40$} \\
    \midrule
    \multicolumn{10}{c}{Small CNN MNIST, 4 layer, $\epsilon_{train}=0.3$} \\
    \midrule
    IBP, IBP Verified & 98\%  & 98\%  & 98\%  & 98\%  & 98\%  & 98\%  & 96\%  & 93\%  & 0\% \\
    IBP, Zero Verified & 98\%  & 98\%  & 98\%  & 98\%  & 98\%  & 98\%  & 97\%  & 93\%  & 0\% \\
    \singlepropzero, IBP Verified & 98\%  & 97\%  & 97\%  & 97\%  & 97\%  & 96\%  & 94\%  & 91\%  & 0\% \\
    \singlepropzero, Zero Verified & 98\%  & 97\%  & 97\%  & 97\%  & 97\%  & 96\%  & 95\%  & 92\%  & 0\% \\
    \midrule
    Method (100 points) & \multicolumn{1}{l}{$\epsilon_{cert}=0$} & \multicolumn{1}{l}{$0.5/255$} & \multicolumn{1}{l}{$1/255$} & \multicolumn{1}{l}{$2/255$} & \multicolumn{1}{l}{$3/255$} & \multicolumn{1}{l}{$5/255$} & \multicolumn{1}{l}{$7/255$} & \multicolumn{1}{l}{$8/255$} & \multicolumn{1}{l}{$9/255$} \\
    \midrule
    \multicolumn{10}{c}{Small CNN CIFAR, 4 layer, $\epsilon_{train}=8/255$} \\
    \midrule
    IBP, IBP Verified & 46\%  & 45\%  & 44\%  & 43\%  & 43\%  & 41\%  & 35\%  & 34\%  & 30\% \\
    IBP, Fast-Lin Verified & 46\%  & 45\%  & 44\%  & 42\%  & 41\%  & \textcolor[rgb]{ 1,  0,  0}{33\%} & \textcolor[rgb]{ 1,  0,  0}{20\%} & \textcolor[rgb]{ 1,  0,  0}{14\%} & \textcolor[rgb]{ 1,  0,  0}{11\%} \\
    IBP, Fast-Lin+IBP Verified & 46\%  & 45\%  & 44\%  & 43\%  & 43\%  & 41\%  & 35\%  & 34\%  & 30\% \\
    \singlepropfastlin, IBP Verified & 41\%  & 39\%  & 39\%  & 39\%  & 37\%  & 33\%  & 30\%  & 30\%  & 30\% \\
    \singlepropfastlin,  Fast-Lin Verified & 41\%  & 39\%  & 39\%  & 39\%  & 37\%  & 33\%  & 30\%  & 29\%  & 26\% \\
    \singlepropfastlin, Fast-Lin+IBP Verified & 41\%  & 39\%  & 39\%  & 39\%  & 37\%  & 33\%  & 30\%  & 30\%  & 30\% \\
    \bottomrule
    \end{tabular}%
  \label{tab:100}%
  }
\end{table*}%

\begin{table*}[htbp]
  \centering
  \caption{Certified accuracies on networks trained with IBP and \singleprop on Small CNN CIFAR trained with $\epsilon_{train} = 8/255$. Certification is performed with IBP and CNN-Cert-Zero certification. Certified accuracies at $\eps=8/255$ are reported on 100 or 200 random test set points. Note that the networks tested are different than in Table~\ref{tab:100}.}
  \adjustbox{max width = \textwidth}{
    \begin{tabular}{p{8.215em}p{4.215em}rr}
    \toprule
    Method & Verifier & \multicolumn{1}{p{4.215em}}{100 points} & \multicolumn{1}{p{4.215em}}{200 points} \\
    \midrule
    IBP   & Zero  & 15\%  & 15.0\% \\
    IBP   & IBP   & 13\%  & 13.0\% \\
    \singlepropfastlin & Zero  & 25\%  & 20.0\% \\
    \singlepropfastlin & IBP   & 24\%  & 19.0\% \\
    \bottomrule
    \end{tabular}%
    }
  \label{tab:100vs200}%
\end{table*}%

\begin{table*}[h!]
    \centering
        \caption{Full test set IBP certified accuracies. \{A\} corresponds to points certifiable by either model, \{B\} corresponds to points in \{A\} correctly classified by both models, and \{C\} is the fraction of points correctly classified by both models in \{A\}. Multiple trials use different random initializations and training batch order.}

    \begin{tabular}{p{15.0em}rrrrr}
    \toprule
    Method & \multicolumn{1}{l}{$\epsilon_{cert}=0$} & \multicolumn{1}{l}{$0.05$} & \multicolumn{1}{l}{$0.10$} & \multicolumn{1}{l}{$0.20$} & \multicolumn{1}{l}{$0.30$} \\
    \midrule
    \multicolumn{6}{c}{Small CNN MNIST, 4 layer, multiple trials, $\epsilon_{train}=0.3$} \\
    \midrule
    IBP mean (5 trials) & 96.58\% & 95.66\% & 94.43\% & 90.82\% & 84.00\% \\
    IBP std (5 trials) & 0.22\% & 0.16\% & 0.07\% & 0.24\% & 0.62\% \\
    SingleProp-zero mean (5 trials) & 94.06\% & 92.85\% & 91.44\% & 87.51\% & 80.65\% \\
    SingleProp-zero std (5 trials) & 0.63\% & 0.81\% & 0.96\% & 1.49\% & 2.36\% \\
    IBP + \singlepropzero \{A\}, Trial 1
    & 97.37\% & 96.65\% & 95.94\% & 93.55\% & 88.83\% \\
    \textbf{Improvement:} 
    Row 5 vs. IBP, Trial 1
    & \textcolor{blue}{+1.16\%} & \textcolor{blue}{+1.28\%} & \textcolor{blue}{+1.59\%} & \textcolor{blue}{+2.62\%} & \textcolor{blue}{+4.01\%}  \\
    \midrule
    \multicolumn{6}{c}{Medium CNN MNIST, 7 layer, $\epsilon_{train}=0.3$} \\
    \midrule
    IBP   & 97.17\% & 96.49\% & 95.59\% & 93.09\% & 88.63\% \\
    \singlepropzero & 97.45\% & 96.55\% & 95.46\% & 92.40\% & 86.05\% \\
    IBP + \singlepropzero  \{A\} & 98.36\% & 97.68\% & 96.97\% & 95.02\% & 90.81\% \\
    \textbf{Improvement:} Row 3 vs. IBP & \textcolor{blue}{+1.19\%} & \textcolor{blue}{+ 1.19\%} & \textcolor{blue}{+1.38\%} & \textcolor{blue}{+1.93\%} & \textcolor{blue}{+2.18\%}  \\
    \midrule
    Method & \multicolumn{1}{l}{$\epsilon_{cert}=0$} & \multicolumn{1}{l}{$2/255$} & \multicolumn{1}{l}{$5/255$} & \multicolumn{1}{l}{$7/255$} & \multicolumn{1}{l}{$8/255$} \\
    \midrule
    \multicolumn{6}{c}{Small CNN CIFAR, 4 layer,  multiple trials, $\epsilon_{train}=8/255$} \\
    \midrule
    IBP mean (3 trials) & 37.85\% & 34.85\% & 30.41\% & 27.60\% & 26.25\% \\
    IBP std (3 trials) & 1.07\% & 0.86\% & 0.55\% & 0.35\% & 0.24\% \\
    \singlepropfastlin mean (3 trials) & 37.29\% & 34.10\% & 28.99\% & 25.92\% & 24.39\% \\
    \singlepropfastlin std (3 trials) & 0.36\% & 0.24\% & 0.35\% & 0.50\% & 0.33\% \\
    IBP + \singlepropfastlin \{A\}, Trial 1 & 48.56\% & 44.90\% & 39.23\% & 35.49\% & 33.79\%  \\
    IBP + \singlepropfastlin \{A\}, Trial 2 & 48.06\% & 44.53\% & 39.05\% & 35.71\% & 34.15\%  \\
    IBP + \singlepropfastlin \{A\}, Trial 3 & 50.48\% & 46.54\% & 40.54\% & 36.68\% & 34.73\%  \\
    IBP + \singlepropfastlin \{B\}, Trial 1 & 26.46\% & 26.31\% & 24.47\% & 24.34\% & 23.81\% \\
    IBP + \singlepropfastlin \{C\}, Trial 1 & 100.00\% & 99.43\% & 92.48\% & 91.99\% & 89.98\% \\
    \textbf{Improvement:} Row 5 vs. IBP, Trial 1 & \textcolor{blue}{+10.51\%} & \textcolor{blue}{+9.82\%} & \textcolor{blue}{+8.41\%} & \textcolor{blue}{+7.52\%} & \textcolor{blue}{+7.22\%} \\
    \textbf{Improvement:} Row 6 vs. IBP, Trial 2 & \textcolor{blue}{+11.60\%} & \textcolor{blue}{+10.83\%} & \textcolor{blue}{+9.41\%} & \textcolor{blue}{+8.57\%} & \textcolor{blue}{+8.16\%}  \\
    \textbf{Improvement:} Row 7 vs. IBP, Trial 3 & \textcolor{blue}{+11.43\%} & \textcolor{blue}{+10.78\%} & \textcolor{blue}{+9.76\%} & \textcolor{blue}{+8.98\%} & \textcolor{blue}{+8.53\%}  \\
    \midrule
    \multicolumn{6}{c}{Large CNN CIFAR, 7 layer, $\epsilon_{train}=8/255$} \\
    \midrule
    IBP   & 46.80\% & 41.15\% & 33.16\% & 28.04\% & 25.68\%  \\
    \singlepropfastlin & 44.36\% & 37.79\% & 29.51\% & 24.28\% & 21.94\% \\
    IBP + \singlepropfastlin \{A\} & 55.87\% & 49.38\% & 40.15\% & 34.41\% & 31.83\%  \\
    \textbf{Improvement:} Row 3 vs. IBP & \textcolor{blue}{+9.07\%} & \textcolor{blue}{+8.23\%} & \textcolor{blue}{+6.99\%} & \textcolor{blue}{+6.37\%} & \textcolor{blue}{+6.15\%} \\
    
    \bottomrule
    \end{tabular}%
    \label{tab:complementary}
\end{table*}

\end{document}